\documentclass{article}
\usepackage{spconf,amsmath,graphicx,hyperref}
\usepackage{tikz}
\usetikzlibrary{positioning, shapes.geometric, arrows.meta}

\usepackage{booktabs} 
\usepackage{float}  
\usepackage{acronym}
\usepackage{graphicx}
\usepackage{eqnarray}
\usepackage{arydshln}
\usepackage{booktabs}
\usepackage{hyperref}
\usepackage{multirow}
\usepackage{xcolor}
\usepackage{nccmath}
\usepackage{amsfonts}
\usepackage{IEEEtrantools}
\usepackage[table]{xcolor}  

\definecolor{lightgreen}{RGB}{200, 230, 201}

\newcommand{\ci}[1]{{\scriptsize $\pm$#1}}
\definecolor{mygreen}{RGB}{34,139,34}   

\newcommand{\secmt}[1]{}

\acrodef{ssl}[SSL]{Self-Supervised Learning}
\acrodef{kd}[KD]{Knowledge Distillation}
\acrodef{sl}[SL]{Supervised Learning}
\acrodef{bds}[BDS]{Balanced Data Sampling}

\title{S-SONDO: Self-Supervised Knowledge Distillation for General Audio Foundation Models}
\name{
    Mohammed Ali El Adlouni$^{\star}$, 
    Aurian Quelennec$^{\star}$, 
    Pierre Chouteau, 
    Geoffroy Peeters, 
    Slim Essid$^{\dagger}$
    \thanks{$^{\star}$ Equal contributions. $^{\dagger}$ S. Essid is now with NVIDIA. The work was performed when he was with Télécom Paris. This work was supported by the Audible project, funded by BPI, and was performed with GENCI-IDRIS resources (Grant 2024-AD011013929R2)} 
}
\address{
    \textit{LTCI}, \textit{Télécom Paris}, \textit{Institut Polytechnique de Paris}, Palaiseau, France 
}
\begin{document}

\maketitle

\begin{abstract}

General audio foundation models have recently achieved remarkable progress, enabling strong performance across diverse tasks. However, state-of-the-art models remain extremely large, often with hundreds of millions of parameters, leading to high inference costs and limited deployability on edge devices. Knowledge distillation is a proven strategy for model compression, but prior work in audio has mostly focused on supervised settings, relying on class logits, intermediate features, or architecture-specific techniques. Such assumptions exclude models that output only embeddings, such as self-supervised or metric-learning models.
We introduce S-SONDO (\textbf{S}elf-\textbf{S}upervised Kn\textbf{O}wledge Distillatio\textbf{N} for General Au\textbf{D}io F\textbf{O}undation Models), the first framework to distill general audio models using only their output embeddings. By avoiding the need for logits or layer-level alignment, S-SONDO is architecture-agnostic and broadly applicable to embedding-based teachers. We demonstrate its effectiveness by distilling two audio foundation models into three efficient students that are up to $61\times$ smaller while retaining up to 96\% of teacher performance. We also provide practical insights on loss choice and clustering-based balanced data sampling. Code is available here
\footnote{\href{https://github.com/MedAliAdlouni/ssondo}{https://github.com/MedAliAdlouni/ssondo}}.

\end{abstract}


\section{Introduction}
\label{sec:intro}

\begin{figure}[t]
    \centering
    \includegraphics[width=0.95\linewidth]{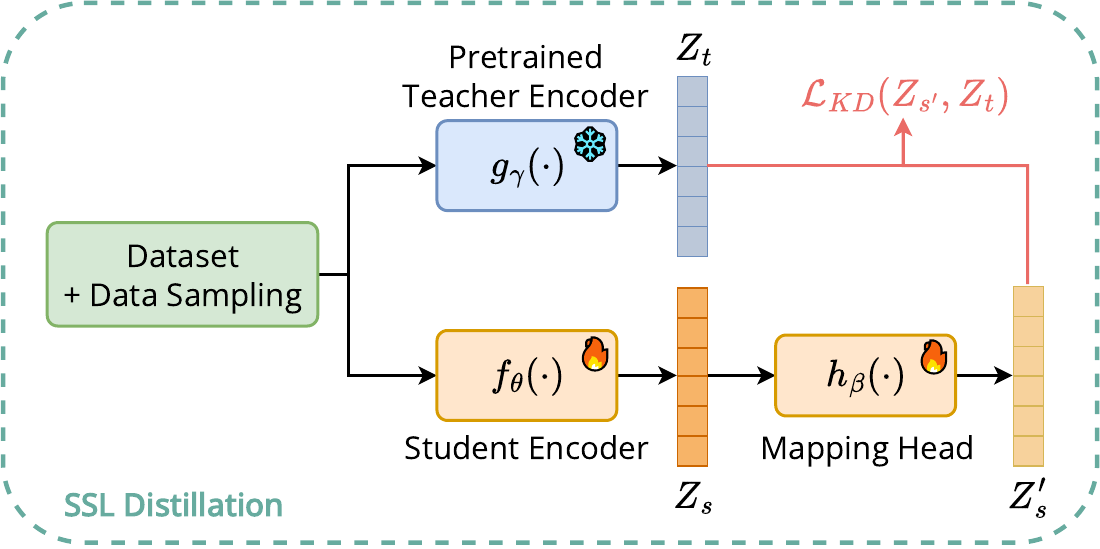} 
    \caption{Overview of the proposed S-SONDO framework. The student embeddings are mapped and aligned with the teacher embeddings in the teacher’s latent space through self-supervised knowledge distillation.}
    \label{fig:ssondo}
    \vspace{-0.5cm}
\end{figure}

\ac{ssl} has recently emerged as a dominant paradigm in representation learning, with remarkable success across speech and general audio. By learning directly from raw data without labels, \ac{ssl} models can leverage vast amounts of unlabeled audio and achieve performance competitive with their supervised counterparts \cite{niizumi23m2d, quelennec2025matpacenhancedmaskedlatent, Ahmed24Asit}. However, these state-of-the-art models typically contain hundreds of millions of parameters, resulting in high inference costs and limiting their deployment in resource-constrained environments such as mobile and embedded devices.

\ac{kd} offers a promising solution by transferring the knowledge of a large teacher model into a smaller student, enabling low-complexity models to achieve competitive performance \cite{hinton, Romero15fitnet, survey}. Yet, the vast majority of these methods rely on \ac{sl} and class logits. Only a few works have considered the case of \ac{ssl} \ac{kd} \cite{Chen_2023_SSSD, Yan20clusterfit, chang_distilhubert_2022, chi2025dicehubert}, and they are applied to the vision and speech domains.
Moreover, they rely either on a specific architecture or do not leverage teacher and student embeddings simultaneously.
To our knowledge, there is currently no work on self-supervised knowledge distillation for general audio.

\noindent\textbf{Contributions} \ \ In this paper, we introduce S-SONDO,
a simple yet effective framework for \ac{ssl} \ac{kd}. 
Inspired by the original \ac{kd} formulation of \cite{hinton}, which only leverages class logits, our approach only relies on teacher and student embeddings.
Prior work \cite{lerch2022featureinformed} aligned student embeddings with the embeddings of a pre-trained teacher while solving a supervised classification task. 
In our proposal, aligning the embeddings becomes a standalone self-supervised training objective, as shown in Fig. \ref{fig:ssondo}. We further enhance \ac{kd} performances by using a \ac{bds} based on the clustering of the teacher's embeddings.


\section{Related Work}

Knowledge distillation (KD) is a widely used method for transferring knowledge from a large pre-trained teacher model to a smaller student model. 
Hinton et al. in \cite{hinton} introduced \ac{kd} to improve the student’s performance by leveraging the teacher’s learned knowledge, matching their output class logits through a Cross Entropy with temperature. 
This method is flexible with respect to the architectures and pre-training of the teacher, but it remains restricted to supervised classification tasks.
FitNet \cite{Romero15fitnet} extended Hinton’s \ac{kd} by aligning intermediate teacher and student representations via Mean Squared Error (MSE), using learned mappings for dimensionality matching. This richer supervision comes at the cost of strong architectural dependencies and continued reliance on class logits. Later work such as \cite{Heo19feat_kd} highlighted the difficulty of designing effective mapping functions. While these approaches show the potential of feature-level alignment, they are unsuitable when only the final embeddings of \ac{ssl} or metric-learning audio models are available.

In the vision domain, \ac{ssl} \ac{kd} has been investigated as an alternative to logit-based supervision, removing the need for class labels. ClusterFit \cite{Yan20clusterfit} trains students with pseudo-labels derived from clustering teacher embeddings, while \cite{Chen_2023_SSSD} extends this idea with layer-wise distillation inspired by FitNet \cite{Romero15fitnet}. However, these approaches have not been applied to general audio models. Neither leverages teacher embeddings directly as a training signal for the student, and the latter is heavily dependent on the student architecture.

For speech, several \ac{ssl} \ac{kd} approaches \cite{chang_distilhubert_2022, chi2025dicehubert} have been proposed to distill specific pre-trained models such as HuBERT \cite{Hsu21HUBERT}. They either align intermediate layer embeddings or train reduced models within HuBERT’s pre-training framework. These methods demonstrate the effectiveness of embedding-based \ac{kd}. Still, they rely on layer-level supervision and require a similar architecture for the student and teacher, leaving open the question of whether aligning only the final embeddings of the teacher and student models provides a sufficient training signal.

Finally, for general audio and music, \ac{kd} has so far been studied exclusively in supervised contexts \cite{SchmidKW24, lerch2022featureinformed, Ding023lerch2}. 
Schmid et al. \cite{schmid2023rnn_kd} adapted the classic \ac{kd} paradigm of \cite{hinton} and explored best practices such as \acf{bds} and mixup. Ding et al. \cite{lerch2022featureinformed, Ding023lerch2} instead aligned embeddings through distance-preserving objectives. Although these works successfully transfer knowledge from general audio teachers, they still depend on classification signals and do not address the open problem of embedding-only distillation for general audio models.

\section{Method}
\label{sec:method}

Figure~\ref{fig:ssondo} illustrates the overall pipeline of our method. Our \ac{kd} strategy follows the formulation introduced in the original \ac{kd} framework \cite{hinton}. We adopt a response-based approach in which the student is trained to align its outputs with those of the teacher. Unlike the standard \ac{kd} setting that matches model logits, our self-supervised \ac{kd} setup instead enforces similarity between the embedding representations produced by the student and the teacher.

\noindent\textbf{Knowledge Distillation}  
We denote the student model by $f_\theta(\cdot)$, parameterized by $\theta$, and the teacher model by $g_\gamma(\cdot)$, parameterized by $\gamma$.
Given an input $X$, the student produces embeddings $Z_s = f_\theta(X) \in \mathbb{R}^{N \times d_s}$, while the teacher produces embeddings $Z_t = g_\gamma(X) \in \mathbb{R}^{N \times d_t}$, where $N$ is the batch size, and $d_s$ and $d_t$ are the dimensions of the student and teacher latent spaces, respectively.

Since $d_s$ and $d_t$ typically differ, we introduce a mapping head $h_\beta(\cdot)$, parameterised by $\beta$, to project the student embedding into the teacher’s latent space. This yields
$Z'_s = h_{\beta}(Z_s) \in \mathbb{R}^{N \times d_t}$.







The projection ensures that the transformed student embeddings $Z'_s$ match the dimensionality of $Z_t$, enabling distillation directly in the teacher’s representation space. This space is semantically structured and well-organised as a result of pre-training, whereas the student latent space is untrained and initially uninformative. Aligning with the teacher’s latent space, therefore, provides the student with a meaningful training signal from the very beginning.

Finally, the projected student embeddings $Z'_s$ and the teacher embeddings $Z_t$ are aligned through a distillation loss $\mathcal{L}_{\text{KD}}(Z'_s, Z_t)$. In this work, we study several candidate formulations for this loss, namely:

\begin{align}
\mathcal{L}_{\text{MSE}}(Z'_s, Z_t) &= \frac{1}{N} \sum_{i=1}^N \lVert Z'^{(i)}_{s} - Z^{(i)}_{t} \rVert_2^2, \\
\mathcal{L}_{\ell_1}(Z'_s, Z_t) &= \frac{1}{N} \sum_{i=1}^N \lVert Z'^{(i)}_{s} - Z^{(i)}_{t} \rVert_1, \\
\mathcal{L}_{\cos}(Z'_s, Z_t) &= \frac{1}{N} \sum_{i=1}^N \bigg( 1 - \frac{\langle Z'^{(i)}_{s}, Z^{(i)}_{t} \rangle}{\lVert Z'^{(i)}_{s} \rVert_2 \, \lVert Z^{(i)}_{t} \rVert_2} \bigg), \\ 
\mathcal{L}_{\text{CLAP}}(Z'_s, Z_t) &= - \frac{0.5}{N} \sum_{i=1}^N\log(l_{Z'_s\rightarrow Z_t} + l_{Z_t\rightarrow Z'_s})^{(i)}\\
\mathcal{L}_{\text{KL}}(Z'_s, Z_t) &= \frac{1}{N} \sum_{i=1}^N \sum_{j=1}^{d_t} 
P_{t}^{(i,j)} \, \log \frac{P_{t}^{(i,j)}}{P_{s}'^{(i,j)}},
\end{align}

\noindent where: $\|\cdot\|_2$ and $\|\cdot\|_1$ are the $\ell_2$ and $\ell_1$ norms; $\langle Z'_{s,i}, Z_{t,i} \rangle$ denotes the dot product; $l_{a\rightarrow b} = diag(\text{softmax}(\langle a, b \rangle))$ for CLAP \cite{Elizalde23clap} loss; $P_t = \text{softmax}(Z_t)$ and $P_s' = \text{softmax}(Z'_s)$ are probability distributions of teacher and student outputs over $d_t$ dimensions. $\mathcal{L}_{\cos}$ is our default loss. 
We evaluate losses commonly used in standard Supervised Learning knowledge distillation, and we aim to determine which one is most suited in our Self-Supervised learning setup. 

\begin{table*}[!ht]
\centering
\vspace{0.5em}

    \resizebox{\textwidth}{!}{%
        \begin{tabular}{l l l | l l l l l l l l }

            
          \textbf{Student} & \textbf{Teacher} & \multirow{2}{*}{\textbf{Size}} & \textbf{OpenMIC} & \textbf{NSynth} & \textbf{GTZAN} & \textbf{MTT} & \textbf{FSD50K} & \textbf{ESC-50} & \textbf{US8K} & \multirow{2}{*}{\textbf{Avg.}} \\
          
          $f_{\theta}(\cdot)$ & $g_{\gamma}(\cdot)$ & & mAP & Acc(\%) & Acc(\%) & mAP & mAP & Acc(\%) & Acc(\%) & \\
        
        \midrule
        
        MobileNetV3 & - & 2.9M & 
        84.5\ci{0.0} & 
        68.0\ci{0.2} & 
        \textbf{87.4\ci{0.1}} & 
        38.7\ci{0.0} & 
        \textbf{49.3\ci{0.0}} & 
        \textbf{92.6\ci{0.3}} & 
        83.7\ci{0.3} & 
        72.0 \\

        \addlinespace[-0.01cm] \hdashline \addlinespace[0.05cm]

        MobileNetV3 & MATPAC++ &  2.9M & 
        \textbf{84.7\ci{0.0}} & 
        \textbf{74.9\ci{0.1}} & 
        85.2\ci{0.0} & 
        \textbf{40.2\ci{0.0}} & 
        48.6\ci{0.1} & 
        91.0\ci{0.1} & 
        \textbf{86.1\ci{0.3}} & 
        \textbf{73.0}\textcolor{mygreen}{\scriptsize (96.4\%)} \\

        MobileNetV3 & M2D &  2.9M & 
        83.1\ci{0.0} & 
        69.5\ci{0.2} & 
        81.4\ci{1.5} & 
        39.5\ci{0.0} & 
        41.4\ci{0.5} &
        85.9\ci{0.1} & 
        83.4\ci{0.4} &
        69.2\textcolor{mygreen}{\scriptsize (93.1\%)}\\

        \midrule
        
        DyMN & - & 
        8.7M & 
        84.3\ci{0.1} & 
        67.5\ci{0.3} & 
        80.8\ci{0.5} & 
        38.8\ci{0.0} &
        47.5\ci{0.0} & 
        91.3\ci{0.1} & 
        83.3\ci{0.1} & 
        70.5 \\

        \addlinespace[-0.01cm] \hdashline \addlinespace[0.05cm]
        DyMN & MATPAC++ & 
        8.7M & 
        \textbf{84.8\ci{0.0}} & 
        \textbf{72.1\ci{0.1}} & 
        \textbf{85.6\ci{0.5}} & 
        \textbf{39.9\ci{0.0}} & 
        \textbf{47.9\ci{0.1}} & 
        \textbf{91.9\ci{0.1}} & 
        \textbf{86.2\ci{0.3}} & 
        \textbf{72.6}\textcolor{mygreen}{\scriptsize (95.9\%)} \\
        
        DyMN & M2D & 
        8.7M & 
        83.1\ci{0.0} & 
        67.3\ci{1.1} & 
        79.2\ci{1.8} & 
        39.1\ci{0.0} & 
        40.6\ci{0.4} & 
        87.4\ci{0.2} & 
        84.1\ci{0.1} & 
        68.7\textcolor{mygreen}{\scriptsize (92.4\%)} \\
        
        \midrule

        ERes2Net & - & 
        1.4M & 
        75.0\ci{0.7} & 
        67.3\ci{0.6} & 
        62.8\ci{3.7} & 
        34.3\ci{0.2} & 
        34.1\ci{0.3} & 
        77.0\ci{1.4} & 
        77.4\ci{0.5} & 
        61.1 \\

        \addlinespace[-0.01cm] \hdashline \addlinespace[0.05cm] 
        ERes2Net & MATPAC++ & 1.4M & 
        \textbf{82.6\ci{0.0}} & 
        \textbf{73.0\ci{0.3}} & 
        \textbf{77.6\ci{0.0}} & 
        \textbf{39.4\ci{0.1}} & 
        \textbf{47.3\ci{0.1}} & 
        \textbf{89.8\ci{0.0}} & 
        \textbf{85.9\ci{0.2}} & 
        \textbf{70.8}\textcolor{mygreen}{\scriptsize (93.5\%)} \\
        
        ERes2Net & M2D & 1.4M & 81.5\ci{0.0} & 
        72.4\ci{0.1} & 
        \textbf{77.6\ci{0.0}} & 
        38.8\ci{0.1} & 
        43.1\ci{0.0} & 
        86.7\ci{0.2} & 
        84.3\ci{0.2} & 
        69.2\textcolor{mygreen}{\scriptsize (93.1\%)} \\

        \midrule

        \textcolor{gray}{-}        & 
        \textcolor{gray}{MATPAC++}   & 
        \textcolor{gray}{86M} & 
        \color{gray}{85.6\ci{0.1}} &
        \color{gray}{76.8\ci{0.2}} &
        \color{gray}{87.6\ci{0.0}} &
        \color{gray}{40.8\ci{0.1}} &
        \color{gray}{56.1\ci{0.1}} &
        \color{gray}{93.1\ci{0.1}} &
        \color{gray}{89.7\ci{0.3}} &
        \color{gray}{75.7} \\ 

        \textcolor{gray}{-}        & 
        \textcolor{gray}{M2D}      &
        \textcolor{gray}{86M}      &
        \color{gray}{84.8\ci{0.0}} &
        \color{gray}{76.2\ci{0.6}} &
        \color{gray}{84.3\ci{1.3}} &
        \color{gray}{40.6\ci{0.1}} &
        \color{gray}{53.4\ci{0.1}} &
        \color{gray}{92.1\ci{0.4}} &
        \color{gray}{88.5\ci{0.3}} &
        \color{gray}{74.3} \\ 

        \end{tabular}
    }
    \vspace{-0.2cm}
    \caption{Downstream evaluation of S-SONDO with 95\% Confidence Intervals (CI). We report the performance of our Knowledge Distillation method across teacher–student combinations. For each student model, supervised training results are reported as a reference (lines where MobileNetV3, DyMN, and ERes2Net have no teacher model). \textbf{Bold} values indicate the best result for each student between supervised and distillation training. \textcolor{gray}{Greyed values} correspond to teacher performance, and \textcolor{mygreen}{green numbers} denote the percentage of teacher performance achieved by the student.}
    \label{tab:res}
\end{table*}

\noindent \textbf{Balanced Data Sampling (BDS)}
In \cite{SchmidKW24}, the authors proposed a label-based sampling strategy to counterbalance infrequent classes and improve supervised \ac{kd}. 
Here, we investigate whether this idea can be adapted to our \ac{ssl} \ac{kd} framework.
Since ground-truth labels are unavailable, we obtain pseudo-labels by clustering the teacher embeddings and use the cluster assignments within the sampling strategy. A sampling weight is assigned to each sample, proportional to the inverse frequency of its cluster $w(i) = \frac{1}{freq(cluster)+100}$, where $i$ is the sample index in the dataset.

\section{Experimental Setup}
\label{sec:exp_setup}
\textbf{Pre-Training setup} 
We use AudioSet \cite{gemmeke2017audioset} as the training dataset, restricted to 10-second clips, yielding 1.8M samples. Each clip is transformed into a log-Mel spectrogram at 32 kHz with a 32 ms window, 16 ms hop, and 128 Mel bins covering 50–16,000 Hz, which serves as input to the student models. For pre-training, we follow the hyperparameters of \cite{schmid2023rnn_kd}. All models are trained for 200 epochs with a batch size of 64 using the Adam optimizer, a base learning rate of $8 \times 10^{-4}$, and a custom learning-rate scheduler. At each epoch, 100,000 clips are sampled without replacement through our unsupervised cluster-based \ac{bds}, with $k=50$ clusters by default.

To illustrate the generalizability of our method, we test it with three different student architectures $f_{\theta}$ and two different teachers $g_{\gamma}$.

\noindent\textbf{Student Model $f_\theta(\cdot)$} \ \ We chose student models with distinct architectural designs and parameter scales.
First, we use MobileNetV3 \cite{howard2019searchingmobilenetv3}, a lightweight architecture shown to be effective in both vision tasks and general audio supervised \ac{kd} \cite{schmid2023rnn_kd}. We adopt its small variant with 2.9M parameters.
Second, we include DynamicMobileNet (DyMN), a modified MobileNetV3 that incorporates dynamic convolutions. DyMN has demonstrated higher efficiency and stronger supervised \ac{kd} performance than MobileNetV3 on general audio tasks, with 8.7M parameters.
Finally, we employ ERes2Net \cite{Chen23ERes2Net}, an efficient model originally introduced for speaker verification. By leveraging multi-scale residual connections and inner-layer feature fusion, it achieves strong performance with as few as 1.4M parameters.

\noindent\textbf{Teacher Model $g_\gamma(\cdot)$} \ \ As teacher models, we adopt two state-of-the-art \ac{ssl} audio foundation models: M2D \cite{niizumi23m2d} and MATPAC++ \cite{quelennec2025matpacenhancedmaskedlatent}. These architectures produce latent audio embeddings rather than class logits, making them well aligned with the S-SONDO method, which relies exclusively on teacher–student representations. Both are Transformer-based models with approximately 86M parameters, orders of magnitude larger than the student models.

\noindent\textbf{Mapping Head $h_\beta(\cdot)$} \ \ The mapping head is a multi-layer perceptron with one hidden layer of dimension 1280.

\noindent\textbf{Downstream Datasets} For downstream evaluation, we follow the protocol used in MATPAC++ \cite{quelennec2025matpacenhancedmaskedlatent} to allow for an easy comparison between the performances of the teacher models and the student ones. The evaluation covers seven audio tagging tasks, among which four music tasks \cite{openmic, gtzan, nsynth, magnatag} and three environmental sound tasks \cite{esc50, us8k, fsd50k}.

\section{Results and discussion}

\noindent \textbf{Main results} \ \ Table~\ref{tab:res} reports the results of S-SONDO across all teacher-student combinations. 
Since this is, to the best of our knowledge, the first \ac{ssl} \ac{kd} framework that relies exclusively on embeddings as the training signal, we compare the performance of $f_{\theta}$ trained using \ac{ssl} \ac{kd} of $g_{\gamma}$ to the ones obtained by training directly $f_{\theta}$ in a supervised way.
This is consistent with prior \ac{kd} work in fully supervised settings.

Two main observations emerge.
First, in 4 out of 6 cases (Avg.=73.0$>$72.0 / 72.6$>$70.5 / 70.8, 69.2 $>$ 61.1), the distilled students outperform their supervised counterparts, confirming the effectiveness of aligning student and teacher embeddings. 
This indicates that the semantic and structural information encoded in the teacher’s latent space is sufficiently rich to guide the student.
Second, across all configurations, students retain at least 92.4\% of the average performance of their respective teachers (green values in the last column). The strongest result is achieved with MATPAC++ as teacher and MobileNetV3 as student, where the distilled model attains 96.4\% of the teacher’s performance while using $\sim$30$\times$ fewer parameters. This setting also yields the highest average score across all distilled student and supervised baseline configurations, with an average of 73.
\begin{table}[ht]
    \centering
    \small
    \resizebox{0.75\linewidth}{!}{%
        \begin{tabular}{l | c c c }
        
         $\mathcal{L}_{\text{KD}}$  & Music Tasks & Env. Tasks & Avg. \\
         \midrule

        Cosine & \textbf{71.0} & \textbf{75.0} & \textbf{72.7} \\
        CLAP & 70.5 & 74.8  & 72.3 \\
        KL Divergence & 65.5 &  67.1 & 66.2\\
        L1 & 69.4 & 60.4 & 65.6 \\
        MSE &  68.2 & 55.3 & 62.7 \\
        
        \end{tabular}
    }
     \vspace{-0.1cm}
     \caption{Loss choice for S-SONDO}
     \label{tab:loss}
\end{table}

\noindent \textbf{Loss choice} \ \ 
In Table 2, we compare the performance of our best model pair (MatPac++ $\rightarrow$ MobileNetV3) with various choices of $\mathcal{L}_{\text{KD}}$. 
Cosine and CLAP yield the best results, with cosine emerging as the most reliable across both music and environmental tasks.
Losses such as L1 and MSE penalise embeddings element-wise, which is suboptimal in latent spaces where semantic information is encoded in relative directions rather than absolute coordinates. Similarly, KL divergence assumes that $Z'_s$ and $Z_t$ are probability distributions, which is not the case for \ac{ssl} embeddings.
CLAP loss can be seen as an extension of cosine loss: it aligns $Z'^{(i)}_s$ and $Z^{(i)}_t$ and forces $Z'^{(i)}_s$ to be dissimilar to all $Z'^{(j)}_s$ with $i \neq j$.
However, in our setting, this contrastive signal is weakened by a limited batch size, 64, and by the lack of control to avoid making embeddings representing similar semantic content be forced apart only because they are not coming from the same sample in the batch.
This explains why CLAP does not surpass cosine in our results. When increasing the batch size (up to 1024), we saw that it started to perform better than the cosine loss, but the scores were less good as with our cosine loss for a batch size of 64.

\begin{figure}[ht]
    \centering
    \includegraphics[width=0.93\linewidth]{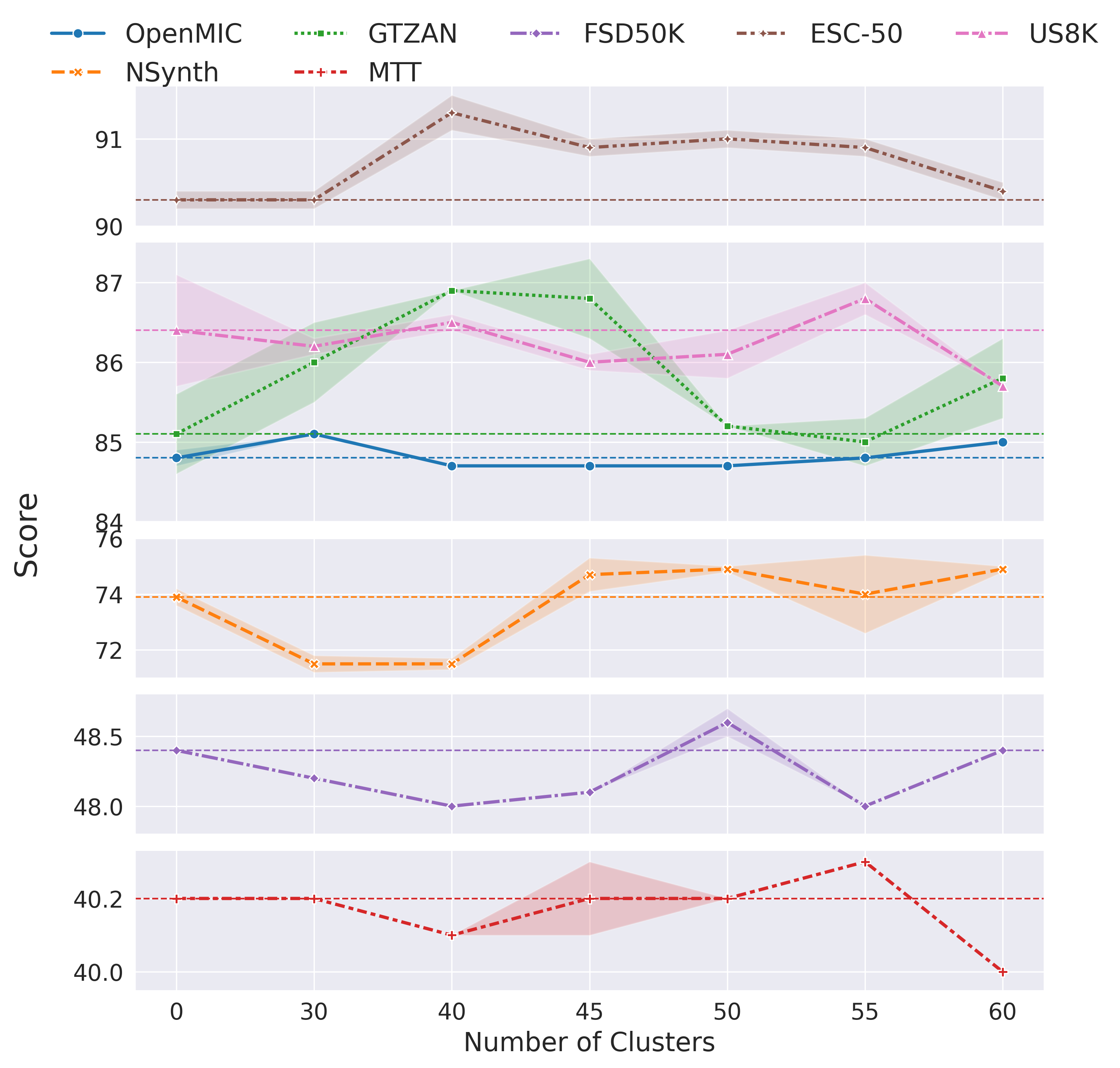}
    \vspace{-0.3cm}
    \caption{Ablation on the number of clusters for the \acl{bds}. The fixed dashed line is the random sampling baseline.}
    \vspace{-0.3cm}
    \label{fig:nclusters}
\end{figure}

\noindent \textbf{\acl{bds}} \ \ We analyze the effect of \ac{bds} using pseudo-labels derived from clustering teacher embeddings. Table~\ref{tab:bds} shows results where $k=50$ is the number of clusters, and Figure~\ref{fig:nclusters} illustrates the impact of varying the number of clusters for (MatPac++ $\rightarrow$ MobileNetV3).
Overall, the effectiveness of \ac{bds} depends on both the teacher and the student. With MATPAC++, two out of three students improve, and ERes2Net fails to converge without it. With M2D, only ERes2Net benefits. This suggests that \ac{bds} is most useful for smaller-capacity students, helping them avoid overfitting or collapse when distilling from large teachers.
The number of clusters also strongly affects performance, but no universal optimum emerges. Tasks with single labels (ESC-50, GTZAN, NSynth, US8K) gain more from clustering than multi-label tasks, likely because k-means struggles to capture complex semantics in multi-label embeddings.

\begin{table}[ht]
    \centering
    \small
    \resizebox{0.75\linewidth}{!}{%
        \begin{tabular}{l | c c }
        
         & w/ BDS & wo/ BDS\\
         \midrule
        
        MATPAC++ $\rightarrow$ MobileNetV3 & \textbf{73.0} & 72.7 \\
        MATPAC++ $\rightarrow$ DyMN & 72.6 & \textbf{72.9} \\
        MATPAC++ $\rightarrow$ ERes2Net & \textbf{70.8} & 44.8 \\
        
        \addlinespace[-0.01cm] \hdashline \addlinespace[0.05cm]
        
        M2D $\rightarrow$ MobileNetV3 & 69.2 & \textbf{69.4} \\
        M2D $\rightarrow$ DyMN & 68.7 & \textbf{69.2} \\
        M2D $\rightarrow$ ERes2Net & \textbf{69.2} & 68.7 \\

        \end{tabular}
    }
     \vspace{-0.1cm}
     \caption{\acf{bds} impact}
     \label{tab:bds}
     \vspace{-0.5cm}
\end{table}



\section{Conclusion}
\label{sec:conclusion}
In this work, we introduced S-SONDO, the first self-supervised knowledge distillation method that aligns student and teacher embeddings without relying on class logits or the model's architecture. 
We evaluated our method using three different students and two teachers. We showed that in 4/6 cases, the performance of our distillations outperforms that of supervised models.
This approach also enables students to retain most of the performance of large teacher models (up to 96.4\%) while reducing model size by up to $61\times$.

Future works will focus on improving balanced data sampling, where pseudo-labels could be made more effective through advanced clustering methods that better capture semantic structure, particularly for multi-label audio tasks. Another promising direction is refining contrastive objectives with improved positive and negative pair selection, combining the strengths of cosine similarity for alignment with contrastive separation to avoid collapse around semantically close negatives.

{\small
\bibliographystyle{IEEEbib}
\bibliography{refs}}

\end{document}